\documentclass{bmcart}

\usepackage[utf8]{inputenc} 
\usepackage{subfiles}
\usepackage{tabularx}
\usepackage{booktabs}
\usepackage{amsmath}
\usepackage{graphicx}
\usepackage{comment} 

\usepackage{soul}
\usepackage{tikz}
\usetikzlibrary{calc}

\usepackage{lineno}

\makeatletter
\newif\if@anonymize

\@anonymizefalse  

\if@anonymize
  \newcommand{\highlight@DoHighlight}{
    \fill [outer sep = -15pt, inner sep = 0pt, color=black]
          ($(begin highlight)+(0,8pt)$) rectangle ($(end highlight)+(0,-3pt)$) ;
  }

  \newcommand{\highlight@BeginHighlight}{
    \coordinate (begin highlight) at (0,0) ;
  }

  \newcommand{\highlight@EndHighlight}{
    \coordinate (end highlight) at (0,0) ;
  }

  \newdimen\highlight@previous
  \newdimen\highlight@current
  \newlength{\item@width}

  \DeclareRobustCommand*\anonymize{%
    \SOUL@setup
    \def\SOUL@preamble{%
      \begin{tikzpicture}[overlay, remember picture]
        \highlight@BeginHighlight
        \highlight@EndHighlight
      \end{tikzpicture}%
    }%
    \def\SOUL@postamble{%
      \begin{tikzpicture}[overlay, remember picture]
        \highlight@EndHighlight
        \highlight@DoHighlight
      \end{tikzpicture}%
    }%
    \def\SOUL@everyhyphen{%
      \discretionary{%
        \SOUL@setkern\SOUL@hyphkern
        \SOUL@sethyphenchar
        \tikz[overlay, remember picture] \highlight@EndHighlight ;%
      }{%
      }{%
        \SOUL@setkern\SOUL@charkern
      }%
    }%
    \def\SOUL@everyexhyphen##1{%
      \SOUL@setkern\SOUL@hyphkern
      \settowidth{\item@width}{##1}%
      \makebox[\item@width]{}%
      \discretionary{%
        \tikz[overlay, remember picture] \highlight@EndHighlight ;%
      }{%
      }{%
        \SOUL@setkern\SOUL@charkern
      }%
    }%
    \def\SOUL@everysyllable{%
      \begin{tikzpicture}[overlay, remember picture]
        \path let \p0 = (begin highlight), \p1 = (0,0) in \pgfextra
          \global\highlight@previous=\y0
          \global\highlight@current =\y1
        \endpgfextra (0,0) ;
        \ifdim\highlight@current < \highlight@previous
          \highlight@DoHighlight
          \highlight@BeginHighlight
        \fi
      \end{tikzpicture}%
      \settowidth{\item@width}{\the\SOUL@syllable}%
      \makebox[\item@width]{}%
      \tikz[overlay, remember picture] \highlight@EndHighlight ;%
    }%
    \SOUL@
  }
\else
  \newcommand{\anonymize}[1]{#1}
\fi
\makeatother

\startlocaldefs
\endlocaldefs

\begin{document}

\begin{frontmatter}

\begin{fmbox}
\dochead{Research paper}
\title{Revising deep learning methods in parking lot occupancy detection}
   
\author[
   noteref={n1},
   addressref={aff1},
   email={}
]{\inits{AM}\fnm{\anonymize{Anastasia}} \snm{\anonymize{Martynova}}}
\author[
   noteref={n1},
   addressref={aff1},
   email={}
]{\inits{MK}\fnm{\anonymize{Mikhail}} \snm{\anonymize{Kuznetsov}}}
\author[
   noteref={n1},
   addressref={aff1},
   email={}
]{\inits{VP}\fnm{\anonymize{Vadim}} \snm{\anonymize{Porvatov}}}
\author[
   addressref={aff1},
   email={eighonet@gmail.com}
]{\inits{VT}\fnm{\anonymize{Vladislav}} \snm{\anonymize{Tishin}}}
\author[
   addressref={aff2},
   email={}
]{\inits{AK}\fnm{\anonymize{Andrey}} \snm{\anonymize{Kuznetsov}}}
\author[
   addressref={aff2},
    corref={aff2},      
   email={\anonymize{semenova.bnl@gmail.com}}
]{\inits{NS}\fnm{\anonymize{Natalia}} \snm{\anonymize{Semenova}}}
\author[
   addressref={aff3},      
   email={}
]{\inits{KK}\fnm{\anonymize{Ksenia}} \snm{\anonymize{Kuznetsova}}}

\address[id=aff1]{%
  \orgname{\anonymize{PJSC Sberbank}},
  \street{\anonymize{Vavilova Street}},
\postcode{\anonymize{117312}}
  \city{\anonymize{Moscow}},
  \cny{\anonymize{Russia}}
}
\address[id=aff2]{%
  \orgname{\anonymize{Artificial Intelligence Research Institute}},
  \street{\anonymize{Nizhniy Susalnyy Lane 5}},
  \postcode{\anonymize{105064}}
  \city{\anonymize{Moscow}},
  \cny{\anonymize{Russia}}
}
\address[id=aff3]{
  \orgname{\anonymize{National University of Science and Technology "MISIS"}},
  \street{\anonymize{Lenin Avenue 4}},                  
  \postcode{\anonymize{119049}}                         
  \city{\anonymize{Moscow}},                            
  \cny{\anonymize{Russia}}                              
}

\begin{artnotes}
\note[id=n1]{Equal contribution}
\end{artnotes}

\end{fmbox}

\begin{abstractbox}

\begin{abstract} 
Parking guidance systems have recently become a popular trend as a part of the smart cities' paradigm of development. The crucial part of such systems is the algorithm allowing drivers to search for available parking lots across regions of interest. The classic approach to this task is based on the application of neural network classifiers to camera records. However, existing systems demonstrate a lack of generalization ability and appropriate testing regarding specific visual conditions. In this study, we extensively evaluate state-of-the-art parking lot occupancy detection algorithms, compare their prediction quality with the recently emerged vision transformers, and propose a new pipeline based on EfficientNet architecture. Performed computational experiments have demonstrated the performance increase in the case of our model, which was evaluated on 5 different datasets.    
\end{abstract}


\end{abstractbox}

\end{frontmatter}

\section*{Introduction}

Extensive motorization has significantly affected the availability of parking lots in large cities. Currently, cruising for a parking lot accounts for 30$\%$ of the traffic flows in certain agglomerations and can take more than 7 minutes on average~\cite{acharya:2018}. In order to address this challenge, various parking guidance systems were developed and subsequently integrated with the traffic management services~\cite{buntic:2012, bong:2008, doulamis2013improving, wada2003development}. Such systems consist of four parts responsible for the data collection, processing, transmission, and final distribution among drivers. The second component of this pipeline frequently involves computer vision algorithms that differentiate the availability of parking lots observed from the system's cameras. While the early approaches frequently involved the Hough transform~\cite{jung2006parking, al2014intelligent}, most of the recent algorithms are based on neural network classifiers, which have repeatedly proven themselves as excellent tools in the various image-related tasks~\cite{zhang:2021, bokhovkin:2019, chen:2015}. 

Despite the wide range of established solutions for parking lot occupancy detection~\cite{huang:2011,amato:2016,nieto:2019}, they still lack the proper generalization ability assessment. Most of the presented approaches rely on the specific setting (e.g., observation angle, weather conditions, illumination intensity, absence of perspective distortions), which can be easily altered in real-world usage scenarios. In this paper, the authors extensively evaluate the current state of the domain on the whole amount of the available datasets and propose a robust solution for parking lot occupancy detection.

We summarize the main contributions of this study as follows:

\begin{enumerate}
    \item We thoroughly studied the state-of-the-art architectures in the area of parking lot occupancy detection, reimplemented main algorithms, and provided an in-depth performance analysis of each considered baseline.
    \item We published a new seasonal parking lot dataset (hereinafter -- SPKL) collected on 5 courtyard parking spaces. In contrast with the previously established datasets, SPKL incorporates wintertime images and multiple types of visual occlusions natural for real-world scenarios. Obtained images were annotated with the help of the special widgets introduced in this study.
    \item We extended available datasets for parking lot occupancy detection with additional visual condition labels and standardised their storage format.
    \item We proposed a new approach to parking lot occupancy detection based on EfficientNet architecture and explored its capabilities regarding multiple configurations. The experiments revealed that our pipeline achieved better metrics for most of the possible evaluation settings.
\end{enumerate}

Corresponding code can be obtained from the anonymous GitHub repository of the project\footnote{https://github.com/Eighonet/parking-research}. The repository content covers all of the considered models and datasets required to reproduce the presented results. 

\section*{Related work}
Classification algorithms for parking lot occupancy detection can be divided into patch-based and intersection-based methods. Models from the first category crop the part of an input image with a specific parking lot and apply a binary classifier to the obtained patch. The intersection-based models rely on the extraction of cars' bounding boxes. In order to make a decision regarding occupancy status, they compare the intersection area of detected vehicles and parking lots with the predefined threshold.   

In this section, we discuss both approaches to parking lot occupancy detection along with the preprocessing techniques that ordinarily increase the prediction quality in similar tasks.


\subsection*{\textbf{Image preprocessing}}

\begin{figure}[!t]

\centering
\includegraphics[width=348pt]{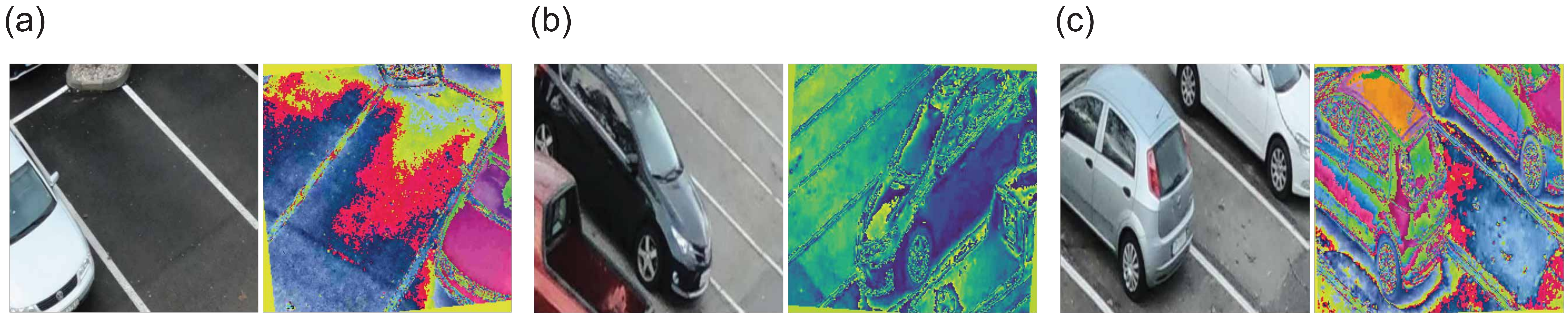}
\caption{
\csentence{Image augmentation examples (simultaneously applied normalization, flipping, rotation, and noise injection).}
The figure demonstrates instances of empty (a) and occupied (b) parking lots, (c) shows the occlusion of the top part of a car and an empty parking lot.
}
\label{rw:augmentations}
\end{figure}

Learning models in computer vision have a tendency to memorize a specific ordering of pixels describing the target objects. By creating several altered versions of an initial image, machine learning algorithms receive more data to learn without collecting and labelling new samples, Figure~\ref{rw:augmentations}. Early studies regarding the convolutional neural networks have shown a positive influence of data augmentation even in the case of the simplest transformations (e.g., image flipping or random cropping~\cite{nanni:2021,shorten:2019}). Specific instances of data augmentation operations can also be applied to the parking lot occupancy detection problem:

\vspace{-8pt}
\subsection*{\textit{Translation}} 
Shifting the input image along the main axis and filling the emerged empty spaces with a constant value or Gaussian noise.
\\

\vspace{-8pt}
\subsection*{\textit{Random cropping}} 
This operation extracts a random rectangular area from the original image. It can be used to reach a similar effect to translation along with a reduction of the image size.

\vspace{-8pt}
\subsection*{\textit{Horizontal flipping}} 
A simple technique that can improve model performance by mirroring an image relative to the vertical axe.

\vspace{-8pt}
\subsection*{\textit{Rotation}} 
An initial image is rotated by a certain number of degrees. This operation also alters the object's position in the frame by transforming the coordinates of its bounding boxes.

\vspace{-8pt}
\subsection*{\textit{Noise injection}} 
This method blends an initial image with the values from the Gaussian distribution, which positively affects the robustness of learning models.

\vspace{8pt}
Described geometric transformations allow algorithms to efficiently avoid positional biases potentially presented in a training sample. In comparison with the other types of augmentation techniques, the mentioned operations do not cause a dramatic increase in memory usage and training time while remaining easy to implement.

\subsection*{\textbf{Patch-based methods for parking lot occupancy detection}}

The common approach to the image classification problem is a sequential application of convolution operations, followed by a feedforward network (FFN). In order to describe the current state of the corresponding research domain, it is convenient to introduce several classic architectures of convolutional neural networks (CNN). 

\vspace{-6pt}
\subsection*{\textit{AlexNet}} 
One of the most well-known CNN pipelines is AlexNet~\cite{krizhevsky:2012}, constructed as the direct extension of the LeNet~\cite{lecun:1998} architecture. This model consists of 5 convolutions and 3 fully connected layers with dropouts. Instead of once applied in LeNet \textit{tanh} activation functions, AlexNet utilizes \textit{ReLU} nonlinearities to achieve a better convergence rate. Further experiments with derivative architectures have led to the appearance of mAlexNet~\cite{amato:2016} with a reduced number of convolutional layers. Despite the substantial decrease in learning parameters, mAlexNet was able to produce better predictions due to the less expressed overfitting tendency.

\vspace{-6pt}
\subsection*{\textit{CarNet}}
As the alternative to the adaption of previously developed architectures, CarNet~\cite{nurullayev:2019} was designed from scratch. It contains dilated convolution layers allowing the model to efficiently capture large-scale features due to the kernel sparsity. The proposed increase of kernel size and applied constraints on the network's depth enabled  CarNet to surpass several previously studied architectures (including AlexNet).

\vspace{-6pt}
\subsection*{\textit{Contrastive occupancy detection}}
Unlike the CarNet model, contrastive occupancy detection~\cite{vu:2019} significantly deviates from the standard CNN pipelines. It was intentionally designed to overcome the inter-object occlusion issues with the help of the spatial transformer network (STN) and contrastive feature extraction network (CFEN). STN aims to produce robust feature representations of image patches with different vehicle sizes, parking displacements, and perspective distortions. To enforce the compactness within the feature space, CFEN utilizes a siamese-style learning approach via two CNNs constrained to have identical weights.

\vspace{-6pt}
\subsection*{\textit{VGG \& ResNet}}
The last generation of parking lot occupancy detection models~\cite{valipour:2016,acharya:2018,singh:2021} continues to use general-purpose CNNs such as VGG~\cite{simonyan:2015} or ResNet~\cite{he:2016}. The first architecture is ordinarily applied in configuration with 19 convolutions as it frequently outperforms the smaller version with 16 layers~\cite{mascarenhas2021,wahyudi2020}. After the appearance of this model, the depth of further state-of-the-art CNNs has demonstrated an increasing trend. Following this trend, the ResNet50 architecture includes 49 convolutional layers and a single fully connected layer. Alongside the depth increase, ResNet overcame a common for VGG vanishing gradient problem via \textit{residual connections}, which establish a relation between two distant layers. 

\vspace{-6pt}
\subsection*{\textit{EfficientNet}}
In contrast to the previously described methods, the EfficientNet~\cite{tan:2019} model family was developed with the help of neural architecture search~\cite{elsken:2019} -- a class of algorithms that automatically construct neural network architectures regarding the desired settings. The main feature of this approach is the scaling method that uniformly changes depth, width, and resolution using the special compound coefficient. 
Despite the absence of parking lot occupancy detection among the previously considered test cases, EfficientNet retains the potential to achieve competitive performance in this task according to its remarkable generalization ability~\cite{ha2020identifying,shah2022dc,wang2022adjusted}.

\vspace{-6pt}
\subsection*{\textit{Vision transformers}}
Besides the widespread CNNs, it is important to mention recently emerged vision transformers~\cite{dosovitskiy:2020} as a completely different approach to image-related tasks. Vision transformers process input by the standard transformer encoder~\cite{vaswani:2017} and construct representations by estimation of relationships among image regions arranged into a linear sequence. In comparison with CNNs, they require significantly larger datasets and demonstrate higher sensitivity to hyperparameter choice. According to these properties, they were never tested regarding the considered task, which encouraged us to study their performance along with conventional methods.
\\

Overall, patch-based models represent the major part of previously applied methods in parking lot occupancy detection. Unveiled capabilities of several state-of-the-art architectures preserve operational space for further experiments: EfficientNet and vision transformers require domain-specific exploration regarding their predictive performance and computational complexity. 


\begin{table}[t!]
    \renewcommand{\arraystretch}{1.5}
    \centering
    \caption{Comparison of discussed methods }
    \begin{tabularx}{360pt}{X|X|X|X}
    \toprule
         Method & training dataset size & computational cost & summary \\ \midrule
         AlexNet, mAlexNet, CarNet &  medium & low & baseline architectures, relatively fast \& easy to implement\\ 
         Contrastive detection (siamese CNN) & low & medium & an alternative approach designed to tackle specific challenges of parking lot occupancy detection\\
         VGG, ResNet & medium & medium & deeper CNNs potentially allowing to  achieve better results \\
         EfficientNet & medium & medium & algorithmically designed neural networks with improved convergence\\
         Vision transformers & large & high &  data-hungry state-of-the-art approach \\
         F-RCNN, RetinaNet & zero-shot 
         & medium & object detection models which do not require specific training on parking lot data\\ \bottomrule
    \end{tabularx}
    \label{rw:methods_summary}
\end{table}

\subsection*{\textbf{Intersection-based methods for parking lot occupancy detection}}


An alternative approach considered in several studies~\cite{nieto:2019,ding:2019} includes object detection models as the first part of a classification pipeline. 

\vspace{-6pt}
\subsection*{\textit{Faster R-CNN}}
One of the classic detection algorithms applicable to parking lot occupancy detection is Faster R-CNN~\cite{ren:2015}. This model was constructed on the basis of the Fast R-CNN~\cite{girshick:2015} architecture, inheriting all of the benefits and drawbacks of two-stage detection algorithms. Faster-RCNN considerably outperformed its earlier counterparts with the help of \textit{region proposal network}, which tangibly simplified the generation of object proposals with respect to Fast R-CNN~\cite{li:2021}.

\vspace{-6pt}
\subsection*{\textit{RetinaNet}}
Another relevant architecture for the considered task is RetinaNet~\cite{lin:2017}, a one-stage object detection model that utilizes a focal loss function to address potential issues with imbalanced samples. Focal loss applies a modulating term to the standard cross-entropy function and focuses the learning process on so-called \textit{hard negative examples}. RetinaNet is an end-to-end pipeline consisting of a backbone neural network paired with two task-specific subnetworks. The backbone network is responsible for the primary extraction of feature maps, which are further utilized by object classification and bounding box regression subnetworks.
\\

The summary of the discussed methods is shown in Table~\ref{rw:methods_summary}.
Compared to patch-based approaches, intersection-based models are underrepresented among the proposed solutions for parking lot occupancy detection. Such a situation encouraged us to explore the mentioned architectures in order to reveal their applied potential in the considered task.

\section*{Preliminaries}
In this section, we define the main concepts for the parking lot occupancy detection problem.

\textbf{Parking lot annotation}. Quadrangle $p_k = \{(x_k{}_i,y_k{}_i)\}_{i=1}^4$ where $x_i \in [0,\, W]$ and $y_i \in [0,\,H]$, $W$ and $H$ denote width and height of the corresponding image.

\textbf{Parking lot patch}. Rectangle $r_k = \{(min(x_k{}_i), \; min(y_k{}_i)),$ $(max(x_k{}_i), \; max(y_k{}_i))\}$ where $(x_k{}_i,y_k{}_i) \in p_k$ (i.e. $r_k$ is a minimum rectangle around $p_k$).

As we mentioned earlier, the parking lot occupancy detection problem can be specified in terms of two different approaches, which imply separate task definitions.

\textbf{Patch-based classification}. For a given set of preventively extracted parking lot patches $R$ from the image set $I$, it is required to classify each $r_k \in R$ as empty or occupied.

\textbf{Intersection-based classification}. For a given image $i \in I$, it is required to locate an arbitrary number of regions of interest and further assign the probability of car presence in each of them. In the second stage, two possible classification heuristics $E_1$ and $E_2$ can be applied to the obtained bounding boxes:
\begin{equation}\label{eq:1}
    \text{E}_1(p_i, b_j) = 
\begin{cases}
    1,& \text{if } \text{IoU}(p_i, b_j)\geq \alpha\\
    0,              & \text{otherwise}
\end{cases}
\end{equation}
where $p_i$ is the annotation of parking lot $i$, $b_j$ is the bounding box of car $j$, $\alpha$ is the adjustable threshold parameter, IoU($p_i,b_j$) is the common area of $p_i$ and $b_j$ divided by their union.  
\begin{equation}\label{eq:2}
    \text{E}_2(p_i, b_j)= 
\begin{cases}
    1,& \text{if } \rho(p_i, b_j)\leq min(p_k)\\
    0,              & \text{otherwise}
\end{cases}
\end{equation}
where $\rho(p_i, b_j)$ is the euclidean distance between polygon centroids of parking lot annotation $p_i$ and car bounding box $b_j$, $min(p_k)$ is the minimum side of annotation quadrangle $p_k$.  

\section*{Data}
\begin{figure}[t!]
\centering
\includegraphics[width=348pt]{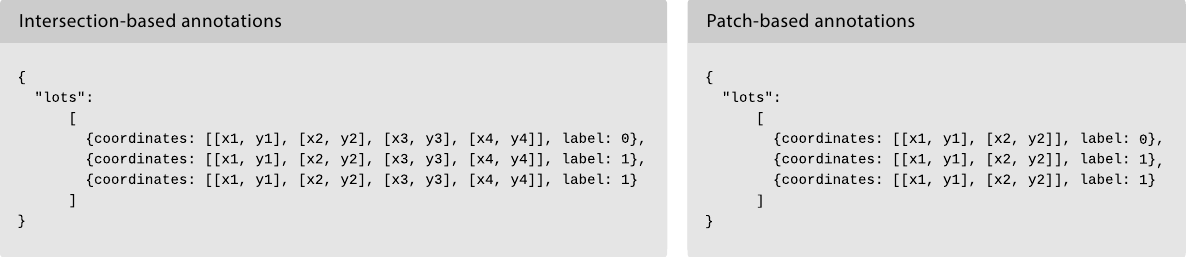}
\caption{
\csentence{Examples of image-level JSON files for two possible types of annotations.} Intersection-based annotations include 4 points per each parking lot defining the corresponding quadrangle, while patch-based annotations contain 2 points of a circumscribing rectangle.
}
\label{data:storage_example}
\end{figure}

In order to properly assess considered models, we intend to utilize the full range of currently available datasets: PKLot~\cite{almedia:2015}, ACMPS~\cite{vu:2019},
CNRPark~\cite{amato:2017}, and ACPDS~\cite{marek:2021}. As these datasets come from different sources, they do not possess a uniform storage format that allows the simultaneous evaluation of intersection-based and patch-based methods. In order to overcome this issue, we standardized parking lot annotations, preserving two versions for each possible approach, Figure~\ref{data:storage_example}.

\vspace{-8pt}
\subsection*{\textit{PKlot}}
Among the mentioned datasets, PKLot is frequently applied during the benchmarking and includes 12417 images organized in 3 weather condition groups: sunny, overcast, and rainy. During the collection process, observation cameras were intentionally located far away from parking lots to provide a clear line of sight. This aspect is the critical limitation of PKLot, as the absence of an appropriate number of occlusions considerably oversimplifies the task, which can cause a negative effect on the system performance in more complex testing environments. 

\vspace{-8pt}
\subsection*{\textit{ACMPS}}
ACMPS is another large dataset consisting of images mostly similar to those presented in PKLot. It includes camera recordings captured from 4 different points of view during the daytime and demonstrates approximately the same occlusion number along with limited variations of lighting conditions. These properties make ACMPS interchangeable with PKLot since both datasets can be considered as mediocre options for edge case testing. 

\vspace{-8pt}
\subsection*{\textit{CNRPark}}
A better alternative to PKLot and ACMPS in the sense of robust evaluation is the CNRPark dataset. Observation cameras were positioned to provide a closer view of parked cars, which induced extra occlusions between them as well as other objects  (e.g.,  trees or lampposts). Similar to PKLot, it contains additional labels allowing to test model performance in different weather conditions. 

\vspace{-8pt}
\subsection*{\textit{ACPDS}}
The smallest of the datasets, ACPDS contains a high number of unique points of view and visual categories. The images were captured by a wide-angle camera, which contributes towards an additional challenge due to the presence of perspective distortions. During the data collection process, authors preserved the distance between a ground level and recording setup with respect to the expected locations of cameras on lampposts. In accordance with its properties, ACPDS can be treated as a useful asset for the evaluation on distorted and occluded parking lots.

\begin{figure}[!t]
\centering
\includegraphics[width=348pt]{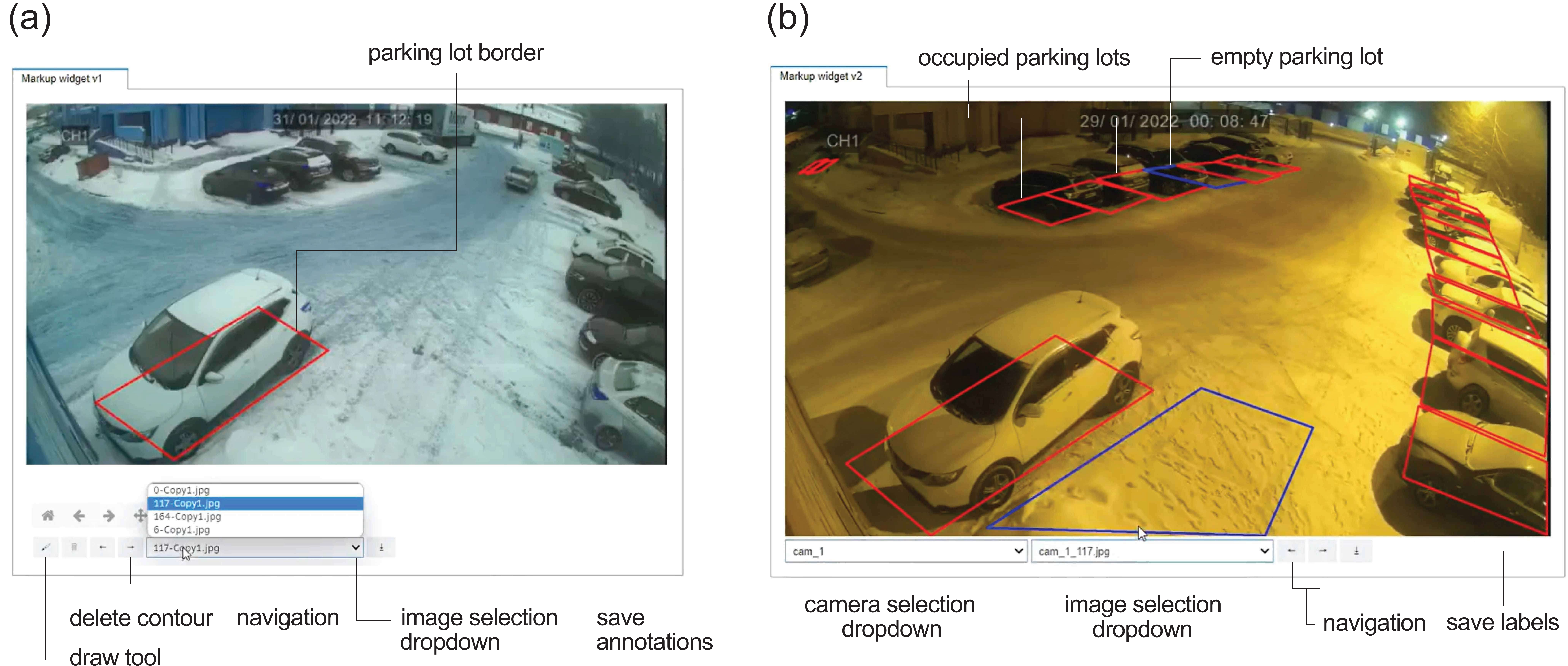}
\caption{
\csentence{Demonstration of widgets' interfaces.}
(a) Position annotation widget enables a potential user to draw and erase quadrangles corresponding to parking lots as well as export the results in the JSON format. (b) The labelling widget requires pre-defined parking lot annotations and provides functionality to assign occupancy status to each entry.}
\label{data:widgets}
\end{figure}

\vspace{-8pt}
\subsection*{\textit{SPKL}}
Despite the variability of data sources, even the cumulative quality of PKLot, ACMPS, CNRPark, and ACPDS  remains far from being satisfactory. None of the datasets contains wintertime images, which remain essential due to specific visual distortions emerging in this season. As the intensive snowfall significantly decreases the visibility of cars, the model trained on currently available datasets is expected to exhibit a performance dropdown. In order to overcome this obstacle, we introduce the SPKL dataset collected in the period between February and March 2022. Among the 1203 entries obtained from 5 parking areas, it includes 440 images with snow varying from a calm environment to a blizzard. The annotations for SPKL were generated with the help of the IPython widgets specially designed to simplify this process, Figure~\ref{data:widgets}.
\\

An additional drawback of the existing datasets comes from insufficient label coverage of presented visual conditions (e.g., fog, night, glare). Since the SPKL dataset required the same annotation procedure, we extended labels associated with each existing dataset up to 11 categories, Table~\ref{data:datasets_table}.

\section*{Method}
As the backbone network of the proposed occupancy detection algorithm, we have selected the relatively lightweight EfficientNet-B0\footnote{https://github.com/lukemelas/EfficientNet-PyTorch} architecture. It includes two types of convolution blocks (denoted as MBConv1 and MBConv6) arranged in 7 sequential modules. 

The starting part of the pipeline is a standard convolution layer with the 3$\times$3 kernel size. The main components of the following MBConv1 and MVConv6 blocks are \textit{depthwise convolution layers}, \textit{swish activation}, and \textit{squeeze} \& \textit{excitation}  subblock. 

\vspace{-8pt}
\subsection*{\textit{Depthwise convolution}} 
In contrast with regular convolutions performed over multiple channels at the same time, their depthwise counterpart splits the input into channels and acts on obtained entries separately. After this procedure, the resulting feature representation is constructed by concatenation of individually convolved channels.

\vspace{-8pt}
\subsection*{\textit{Swish activation}}
The concept of swish activation emerged from the successful attempt to replace classic ReLU nonlinearity, possessing problems with the processing of negative inputs:
 \begin{equation}
    \operatorname{swish}(x)=x\circ\sigma(x),
\end{equation}
where $x$ corresponds to an input matrix, $\sigma$ is a sigmoid activation function, and $\circ$ is the Hadamard product. According to previously established experiments, the swish activation performs better than its direct alternatives such as Leaky ReLU, ELU, and SELU~\cite{ramachandran2017searching}. 

\vspace{-8pt}
\subsection*{\textit{Squeeze \& excitation subblock}} 
When regular CNN produces a feature representation, it does not differentiate the input channels in the sense of their weightage. S\&E blocks overcome this problem with the help of a gating subnetwork designed in a residual manner: along with the base convolution, it learns adaptive weights from the same input and applies them to the resulting feature representation. The crucial property of this design is the opportunity to learn introduced coefficients end-to-end, which allows to implement S\&E blocks in any existing CNN easily.
\\

\begin{table*}[!t]

\renewcommand{\arraystretch}{1.5}

\centering
\caption{Number of images in the extended datasets regarding different types of visual conditions}
\smallskip

\begin{tabularx}{360px}{X|X|X|X|X|X|X}
\toprule
\multicolumn{2}{c|}{property $\backslash$ dataset} & \multicolumn{1}{l|}{PKLot} & \multicolumn{1}{l|}{CNRPark} & \multicolumn{1}{l|}{ACPDS} &
\multicolumn{1}{l|}{ACMPS} & \multicolumn{1}{l}{SPKL}\\
\midrule
\multicolumn{2}{c|}{Total images} & 12417 & 3119 & 293 & 13126 & 1203\\
\midrule
\multicolumn{2}{c|}{Sunny} & 6913 & 1075 & 134 & 2231 & 187\\
\multicolumn{2}{c|}{Overcast} & 4162 & 915 & 132 & 312 & 76\\
\multicolumn{2}{c|}{Rainy} & 1342 & 576 & 13 & 275 & 87\\
\multicolumn{2}{c|}{Winter} & 0 & 0 & 0 & 0 & 440\\
\multicolumn{2}{c|}{Fog} & 0 & 33 & 34 & 0 & 0\\
\multicolumn{2}{c|}{Glare} & 0 & 26 & 1 & 20 & 27\\
\multicolumn{2}{c|}{Night} & 0 & 27 & 13 & 4 & 507\\
\multicolumn{2}{c|}{Infrared} & 0 & 0 & 0 & 4 & 495\\
\multicolumn{2}{c|}{Occlusion (car)} & 7 & 54 & 87 & 0 & 1203\\
\multicolumn{2}{c|}{Occlusion (tree)} & 10 & 3119 & 35 & 0 & 0\\
\multicolumn{2}{c|}{Distortion} & 0 & 0 & 293 & 0 & 400\\
\bottomrule
\end{tabularx}
\label{data:datasets_table}
\end{table*}

\begin{figure}[!t]
\centering
\includegraphics[width=348pt]{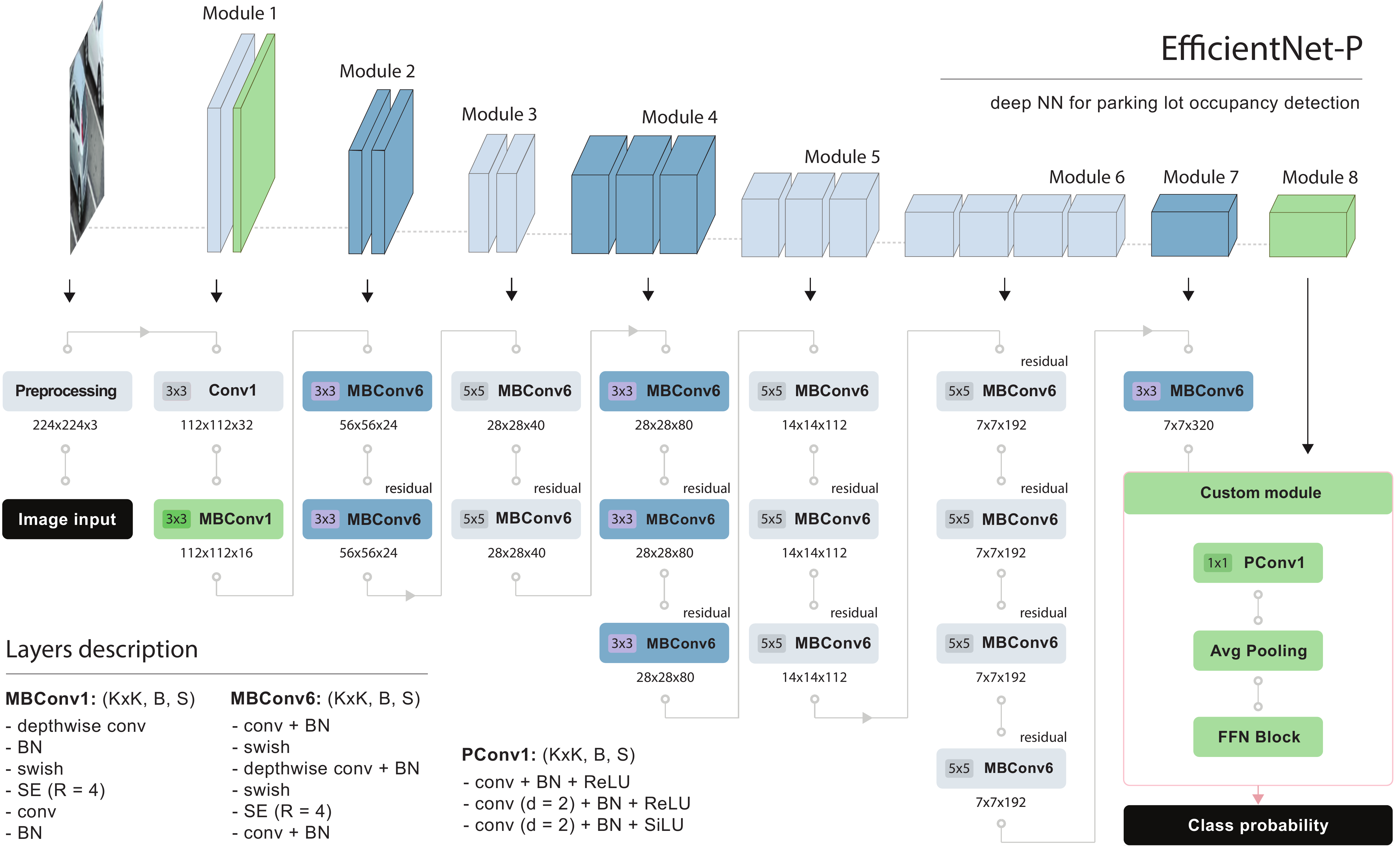}
\caption{
\csentence{The final version of the EfficientNet-P model.}
The custom module follows the first seven blocks of the original EfficientNet-B0.
}
\label{method:pipeline}
\end{figure}

The final version of the proposed EfficientNet-P model is represented in Figure~\ref{method:pipeline}. The MBConv1 block first applies depthwise convolution to the input, followed by swish activation and S\&E block. Depending on the module, its last convolution can be extended by residual connection. In contrast with MBConv1, MBConv6 starts with regular convolution paired with batch normalization. After following swish activation, it applies depthwise convolution and the next S\&E block. The last module of the pipeline is the original part constructed during the fine-tuning of EfficientNet-P. It includes subsequently applied dilated convolutions with ReLU and SiLU activation functions.

\section*{Results}
In this section, we discuss the results obtained during the conducted computational experiments.

\subsection*{\textbf{Experimental setup}}

In order to comprehensively assess the intersection-based approaches, we have selected 3 different backbones for the Faster-RCNN and RetinaNet detectors:
ResNet50, MobileNet, and VGG-19. Considered patch-based methods include variations of the classic VGG, AlexNet and ResNet models, as well as domain-specific neural networks (e.g., CarNet, contrastive occupancy detection). Additionally, we have performed experiments regarding ViT, DeiT, and PiT architectures of vision transformers to demonstrate their capabilities regarding parking lot occupancy detection.

The final configuration of data augmentations includes resizing of images to $224 \times 224$, random rotation on $\pm$15 degrees, horizontal flipping, and channel-wise normalization with $\mu=(0.485, 0.456, 0.406)$ and $\sigma=(0.229, 0.224, 0.225)$. The models were trained on 10 Tesla V100 GPUs and 920 Gb of RAM with the Adam~\cite{kingma2014adam} optimizer. 

According to the class imbalance presented in several datasets, we utilized F1-score as the main evaluation metric:
 \begin{equation}
    \text{F1}=2\cdot\frac{\operatorname{Precision} \cdot \operatorname{Recall}}{\operatorname{Precision} + \operatorname{Recall}},
\end{equation} 
where \textit{Precision} denotes the fraction of true positives among the predicted positive labels, and \textit{Recall} corresponds to the fraction of positive labels that were correctly predicted. 

Established results were obtained from a 5-fold cross-validation with a split ratio 6:1:3 unless otherwise specified. In the main part of the paper, we trained models on the merged data and then evaluated on each dataset. On top of that, we also performed separate training with the further evaluation on corresponding test part, Supplementary 1.



\subsection*{\textbf{Intersection-based models}}

Performed computational experiments with intersection-based methods revealed high variability of metric values across different types of backbone networks, Table~\ref{results_table:od}. In most of the cases, the (\ref{eq:1}) heuristic demonstrates considerably better results than (\ref{eq:2}), excluding the ACPDS dataset. This can be explained by the fact that the (\ref{eq:1}) function relies on overlapping areas, while its counterpart takes into account the shape of the bounding box. According to this feature, (\ref{eq:2}) has a pronounced tendency to provide false negative results for elongated bounding boxes frequently presented in images with elevated observation angles. 

Unlike the other cases, the presence of large-sized annotations and perspective distortions negatively affected (1) predictions. Such an outcome enabled the models based on (2) to considerably outperform other intersection-based methods on SPKL, emphasizing the potential application domain for each considered heuristic.

\begin{table}[!t]
        \renewcommand{\arraystretch}{1.5}
    \centering
    \caption{F1-score for intersection-based models}
    \smallskip
    \begin{tabular}{ll|l|l|l|l|l|l}
    \toprule
        Evaluation & OD & Backbone & ACMPS & ACPDS & CNRPark & PKLot & SPKL \\ 
        \midrule
        E$_1$ & FasterRCNN & ResNet & \textbf{0.7527} & 0.7972 & 0.9629 & \textbf{0.9031} & \textbf{0.8848} \\ 
        ~ & ~ & MobileNet & 0.7251 & 0.7636 & 0.9256 & 0.8952 & 0.6195 \\ 
        ~ & ~ & VGG-19 & 0.7282 & 0.7228 & 0.9185 & 0.8951 & 0.8596 \\ 
        ~ & RetinaNet & ResNet & 0.7384 & 0.7856 & \textbf{0.9683} & 0.8313 & 0.8304 \\ 
        ~ & ~ & MobileNet & 0.7069 & 0.7552 & 0.9234 & 0.8125 & 0.7427 \\ 
        ~ & ~ & VGG-19 & 0.7194 & 0.7234 & 0.9126 & 0.7823 & 0.7581 \\ 
        \midrule
        E$_2$ & FasterRCNN & ResNet & 0.5815 & 0.7963 & 0.4728 & 0.7598 & 0.4727 \\ 
        ~ & ~ & MobileNet & 0.5456 & 0.8013 & 0.4253 & 0.7285 & 0.2906 \\ 
        ~ & ~ & VGG-19 & 0.5298 & 0.7857 & 0.4272 & 0.7347 & 0.4792 \\
        ~ & RetinaNet & ResNet & 0.5115 & \textbf{0.9586} & 0.4926 & 0.7736 & 0.7826 \\ 
        ~ & ~ & MobileNet & 0.5267 & 0.9283 & 0.4163 & 0.7561 & 0.7953 \\ 
        ~ & ~ & VGG-19 & 0.4935 & 0.9126 & 0.4012 & 0.7355 & 0.7924 \\ 
        \bottomrule
    \end{tabular}
    \label{results_table:od}
\end{table}

\subsection*{\textbf{Patch-based models: Vision Transformers}}

As the classic ViT architecture showed unstable performance on relatively small datasets, we applied \textit{shifted patch tokenization} (SPT) and \textit{locality self-attention} (LSA) methods to overcome issues with the lack of training data, Table~\ref{results_table:vt}. The idea behind SPT is the incorporation of several translational augmentations in specific parts of the model, while LSA suppresses diagonal components of the Key-Query matrix, enhancing the quality of attention distribution. 

The inefficiency of SPT and LSA encouraged us to utilize the pre-trained version of ViT (ImageNet) along with several alternative architectures handling input differently. The first considered option was the DeiT model, which is based on the teacher-student distillation strategy. The \textit{distillation token} helps the model to learn from the teacher's output by interacting with the standard \textit{class} and \textit{patch tokens} via self-attention layers. Despite its optimized architecture, the DeiT model did not surpass the metric values of the standard ViT. 

The second tested model was PiT, which introduced a token downsampling procedure to the domain of vision transformers. After the standard ViT encoder, PiT applies a pooling layer based on depthwise convolution to achieve a spatial reduction of propagated tokens. In addition to the considerable decrease of computational effort, such an extension enabled PiT to obtain the best results on the SPKL dataset among patch-based approaches.

\begin{table}[!t]
        \renewcommand{\arraystretch}{1.5}
    
    \centering
    \caption{F1-score for patch-based methods (vision transformer classifiers)}
    \smallskip
    \begin{tabularx}{355px}{XX|X|X|X|X|X}
        \toprule
        \multicolumn{2}{l|}{} & ACMPS & ACPDS & CNRPark & PKLot & SPKL \\ 
        \midrule
        \multicolumn{2}{l|}{ViT} & 0.9789 & 0.8152 & \textbf{0.9193} & \textbf{0.9966} & 0.7335 \\ 
        \multicolumn{2}{l|}{ViT (SPT LSA)} & 0.9339 & 0.8055 & 0.9179 & 0.9906 & 0.7412 \\ 
        \multicolumn{2}{l|}{ViT (pre-trained)} & \textbf{0.9856} & \textbf{0.8263} & 0.9083 & 0.9856 & 0.6600 \\
        \multicolumn{2}{l|}{DeiT (pre-trained)} & 0.9556 & 0.8209 & 0.8902 & 0.9928 & 0.7010 \\ 
        \multicolumn{2}{l|}{PiT (pre-trained)} & 0.9334 & 0.7122 & 0.9084 & 0.9818  & \textbf{0.7455} \\ 
        \bottomrule
    \end{tabularx}
    \label{results_table:vt}
\end{table}

\begin{table}[!t]
        \renewcommand{\arraystretch}{1.5}
    
    \centering
    \caption{F1-score for patch-based methods (convlutional neural network classifiers)}
    \smallskip
    \begin{tabularx}{355px}{XX|X|X|X|X|X}
        \toprule
        \multicolumn{2}{l|}{} & ACMPS & ACPDS & CNRPark & PKLot & SPKL\\
        \midrule
        \multicolumn{2}{l|}{ResNet50} & 0.9379 & 0.8407 & 0.9380 & 0.9926 & 0.6674 \\
        \multicolumn{2}{l|}{MobileNet} & 0.9971 & \textbf{0.9343} & 0.9663 & 0.9991 & 0.6910 \\ 
        \multicolumn{2}{l|}{CarNet} & 0.9882 & 0.8377 & 0.9332 & 0.9984 & 0.7131 \\ 
        \multicolumn{2}{l|}{AlexNet} & 0.9914 & 0.8820 & 0.9555 & 0.9990 & 0.7132 \\ 
        \multicolumn{2}{l|}{mAlexNet} & 0.9813 & 0.8577 & 0.9176 & 0.9959 & 0.6937 \\ 
        \multicolumn{2}{l|}{VGG-16} & 0.9923 & 0.8936 & 0.9496 & 0.9985 & 0.5700 \\ 
        \multicolumn{2}{l|}{VGG-19} & 0.9877 & 0.9152 & 0.9629 & 0.9988 & 0.6801 \\ 
        \multicolumn{2}{l|}{CFEN} & 0.9829 & 0.8302 & 0.8482 & 0.9966 & 0.5367 \\ 
        \midrule
        \multicolumn{2}{l|}{EfficientNet-B0} & 0.9950 & 0.9259 & 0.9572 & 0.9993 & 0.7330 \\ 
        \multicolumn{2}{l|}{EfficientNet-P} & \textbf{0.9982} & 0.9125 & \textbf{0.9683} & \textbf{0.9995} & \textbf{0.7393} \\ 
        \bottomrule
    \end{tabularx}
    \label{results_table:cnn}
\end{table}

\subsection*{\textbf{Patch-based models: Convolutional Neural Networks}}

The results provided by conventional CNNs emphasize the MobileNet and EfficentNet architectures as the most suitable options for parking lot occupancy detection, Table~\ref{results_table:cnn}. These models had reached the comparative F1-score while the domain-specific architectures (e.g., CarNet or contrastive occupancy detection) performed tangibly worse on each of the considered datasets. Among the classic architectures, VGG demonstrated a considerable predictive performance, yielding only to the 2 best models. Another aspect of high importance is the difference between CNNs and vision transformers' results on the SPKL dataset: neither of the top-ranked models did manage to reach better metrics than those obtained by PiT. 

In general, patch-based models achieved better F1-score values on the largest datasets, while intersection-based approaches better processed the minor ones like ACPDS and SPKL. Despite the certain differences in metric values, the performance gap still allows us to prioritize patch-based models: the success of default EfficientNet-B0 preserves the space for further experiments with configurations of the last layers. 

\subsection*{\textbf{Patch-based models: EfficientNet-P}}

The final configuration of EfficientNet-P was designed during the experiments with several architectures altering the last module of EfficientNet-B0. Noteworthy gain in performance was observed for the models extending regular convolutions by additional residual connections (following the design of the residual version of MBConv6) and dilated filters. Among these highlighted approaches, the last one performed more stable on the validation data: implementation of 2 dilated convolutions on top of a single regular one increased the metrics for 4 datasets out of 5. The idea of a dilated module replacing the end part of EfficientNet-B0 was inspired by experiments with the CarNet architecture which, however, did not yield great results by itself. The crucial drawback of this model appeared to be its limited depth which could not be fully compensated by the benefits of kernel sparsity in this task.

\vspace{-8pt}
\subsection*{\textit{Visual categories analysis}}
\begin{figure}[!t]
\centering
\includegraphics[width=348pt]{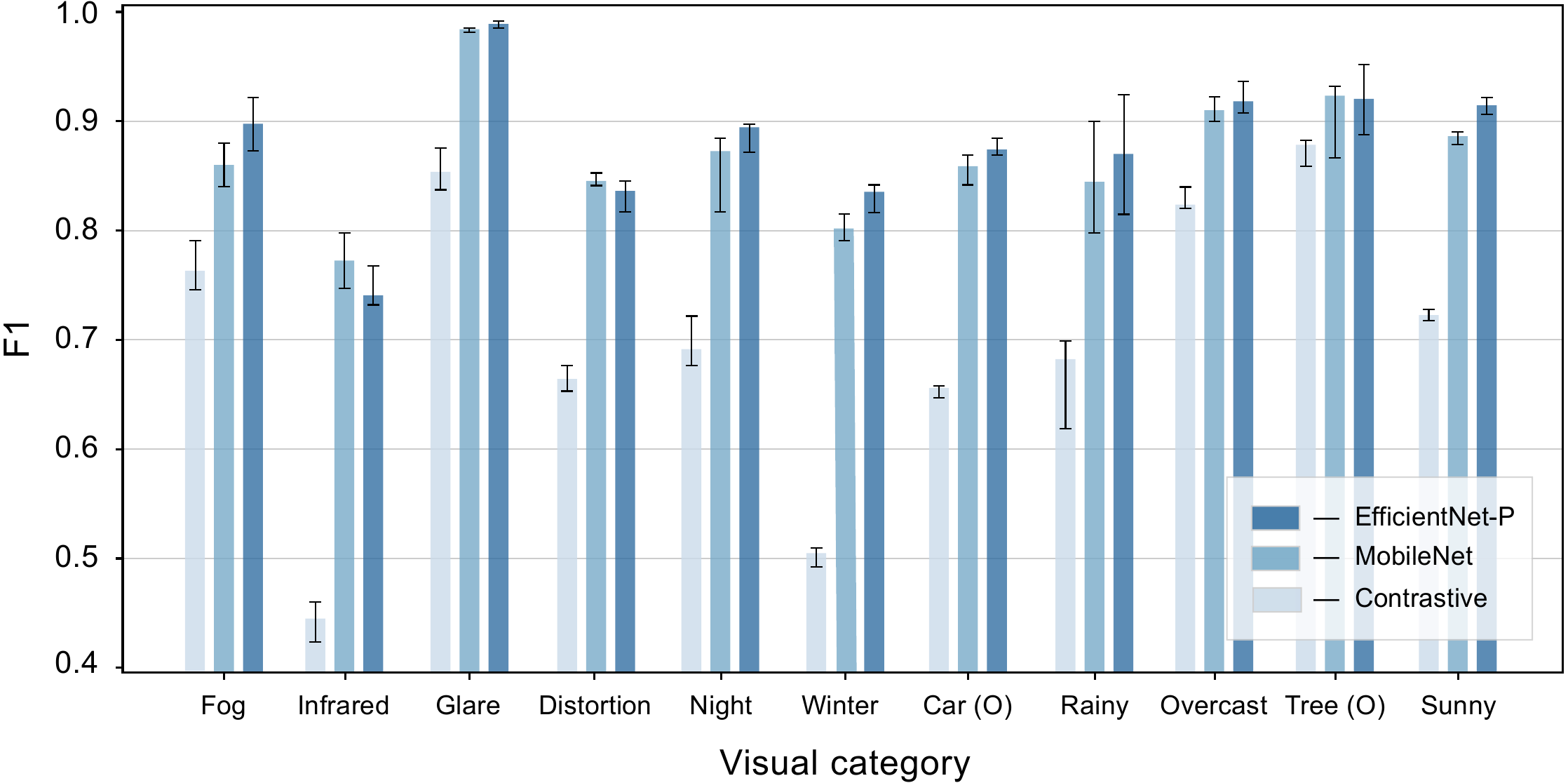}
\caption{
\csentence{Comparison of F1-scores for several considered models for various vision conditions.}
The bars are organized in ascending order with respect to the size of the categories; whiskers are based on the $Q1 - 1.5 \cdot \operatorname{IQR}$ and $Q3 + 1.5 \cdot \operatorname{IQR}$ values from 8 different splits.}
\label{results:error_barcharts}
\end{figure}

To deepen the performed evaluation, it is important to explore the predictive quality of the models on the separate visual categories. We selected contrastive occupation detection due to design properties and MobileNet as the competitive counterpart of EfficientNet-P, Figure~\ref{results:error_barcharts}. It is noteworthy that for most of the categories EfficientNet-P demonstrated better or the same prediction quality with respect to both compared networks. The obtained results indicate that considered models have difficulties in similar cases: the challenging visual conditions are infrared records, winter weather, and inter-car occlusions. Such an outcome depicts the importance of performed dataset extension since the metrics dropdown was observed precisely for newly labelled entries. 

Occlusions induced a significant number of problems across all of the considered datasets. Regardless of the position of objects in a frame, parking lots are frequently overlapped by the top part of the adjacent cars, obstructing the view and confusing a model. This tendency is especially pronounced in the images received through wide-angle cameras (e.g., ACPDS, SPKL) as well as from viewpoints located near the ground (CNRPark). Supporting this claim, Table~\ref{results_table:cnn} demonstrates the extent of the influence of these properties: the associated F1-score on the mentioned datasets is considerably lower than on the others.

Winter images provided with the SPKL dataset naturally hinder the recognition due to large amounts of snow, covering both the depicted cars and surrounding areas. Compared to the other categories, this one not only disrupts the visual integrity of the cars but makes them blend with a background. This unique influence is also reflected in low values of evaluation metrics, similar to those achieved on distorted images, Figure~\ref{results:error_barcharts}. In our experiments, we observed a negative correlation between snow intensity and quality of recognition, which emphasizes the necessity of separate evaluation of this category in future studies. Finally, since the primary source of the infrared images is the SPKL dataset, a significant part of them also belongs to the winter category. This aspect raises the prediction complexity for the corresponding category due to the combined greyscale effect and abundant snow presence.

Variance in each category appeared to be approximately the same for all of the tested models. This characteristic does not have explicit dependency on the size of image groups as expected. It achieves remarkably low values for the sunny weather (the largest group) and glared images (one of the smallest): the semantic proximity of these categories suggests that they have the same level of difficulty for learning models. Rainy weather and tree occlusions induced the largest difference among the metrics corresponding to each category. While the effect of trees can be attributed to the visual obstacles violating visual integrity of the classified objects, results on the second category are not so straightforward to interpret. We assume that observed variance is a product of an illumination shift along with the changes in the display of the ground (e.g., wet asphalt, objects reflections in puddles) which contributes towards both relatively low mean score and worse convergence dynamics.




\vspace{-8pt}
\subsection*{\textit{Augmentation influence}}

Limited number of images in certain datasets necessitates the usage of previously introduced image augmentations (horizontal flipping, rotation, noise injection) to ensure proper generalization ability for separate training. The impact of these augmentations was measured on the two smallest datasets: SPKL and ACPDS. We utilized the Mann-Whitney U test with $\alpha = 0.05$ assuming that the null hypothesis is an absence of difference between metrics in augmented and non-augmented cases. Test data included two samples of 10 F1-scores which were obtained during the evaluation of separately trained models. The resulting p-values 0.82 and 0.046 indicate that the null hypothesis can be rejected only for ACPDS implying that applied augmentations have non-random impact on the model performance for certain data distributions.

\subsection*{\textbf{Further improvements}}

\begin{figure}[!t]
\centering
\includegraphics[width=348pt]{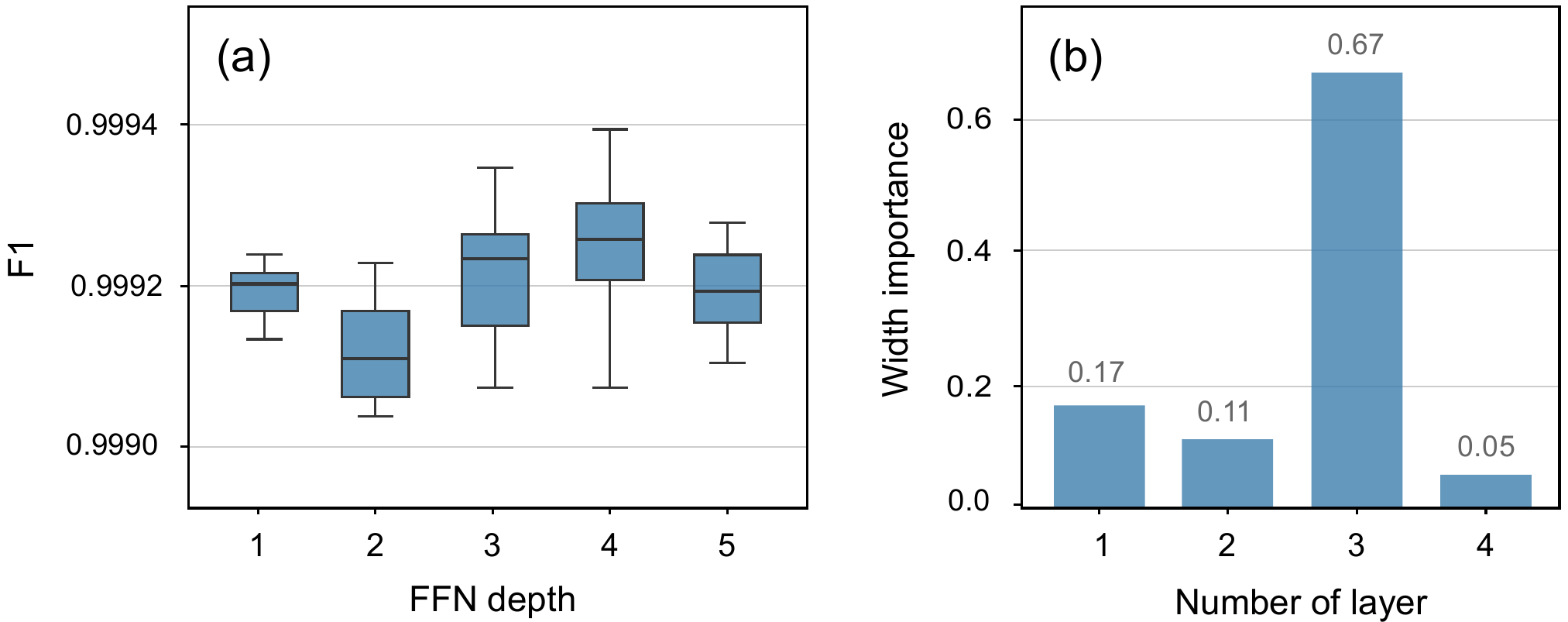}
\caption{
\csentence{Ablation study of EfficientNet-P.} (a) F1-scores of 
 the model with different depths of FFN; (b) Width value importance for each layer in the final configuration.
}
\label{results:ablation_study}
\end{figure}

The base EfficientNet-P model demonstrates better predictions both for the minor and major visual categories. In order to achieve such results, we conducted an ablation study regarding the configuration of the last FFN block, Figure~\ref{results:ablation_study}. The final configuration of FFN includes 4 layers with corresponding widths (558, 191, 49, 7). Additionally, we explored several techniques which had shown a positive influence on the EfficientNet performance for similar tasks~\cite{labatie2021proxy}.

\begin{figure}[!t]
\centering
\includegraphics[width=348pt]{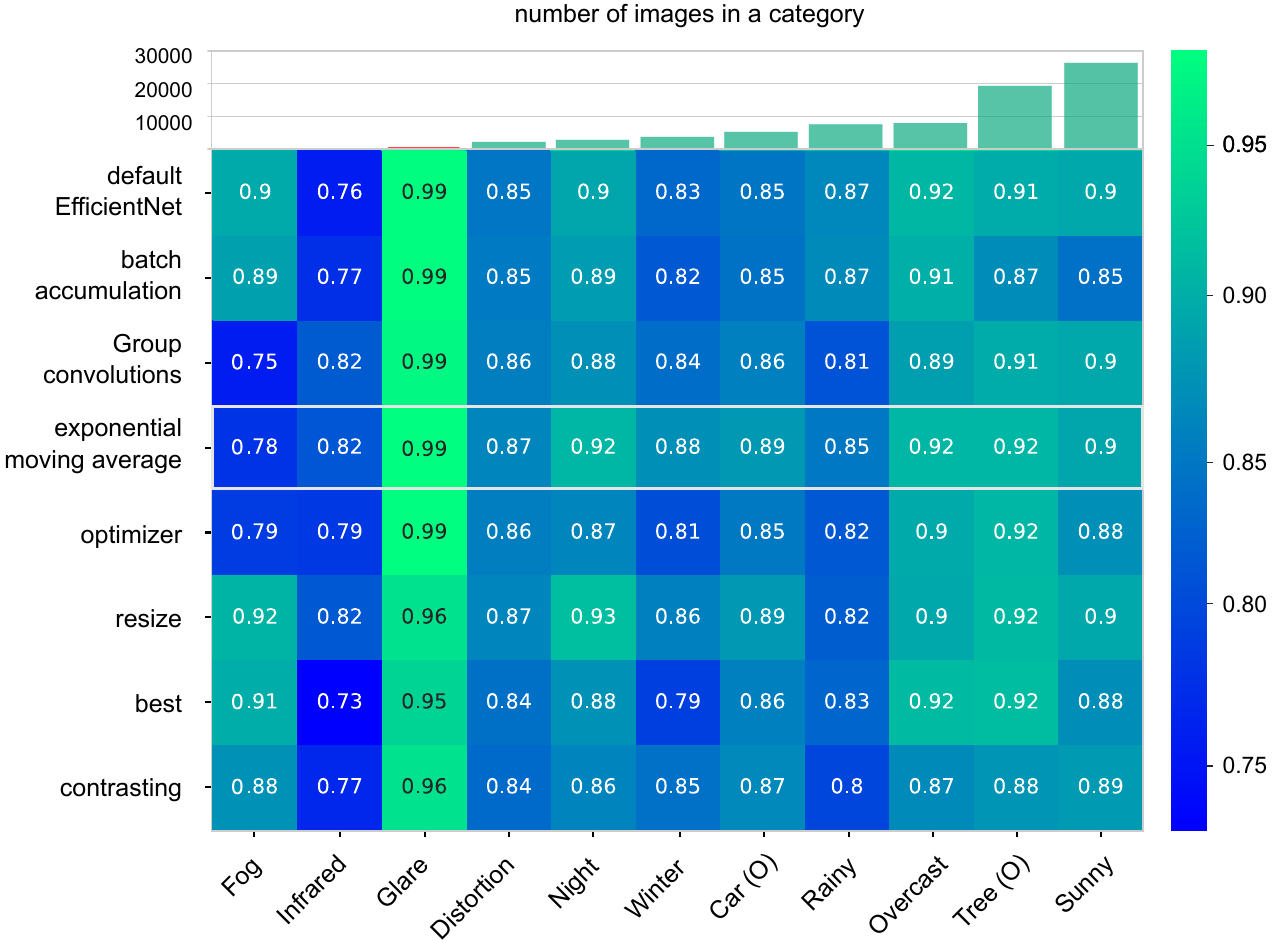}
\caption{
\csentence{Analysis of EfficientNet-P improvements.}
Dependency between F1-score on different categories and chosen modification of EfficientNet-P.
}
\label{results:heuristics}
\end{figure}

\vspace{-8pt}
\subsection*{\textit{Grouped convolutions}}
Grouped convolutions correspond to the use of multiple kernels per convolutional layer. This addition helps the model to capture a greater number of high-level features. 

\vspace{-8pt}
\subsection*{\textit{Resize}}
Experiments with changing the resolution of input images revealed that 0.5 input size reduction could lead to better prediction quality.

\vspace{-8pt}
\subsection*{\textit{Optimizer}}
Replacement of standard Adam optimizer to RMSProp with the learning rate = 0.0001.

\vspace{-8pt}
\subsection*{\textit{Exponential Moving Average}}
Averaging of learning parameters before and after the optimization step enables avoiding the sudden changes in model weights during the training process.

\vspace{-8pt}
\subsection*{\textit{Batch accumulation}}
This technique corresponds to the accumulation of gradients and spreading them back every few batches.

\vspace{-8pt}
\subsection*{\textit{Batch-independent normalization}}
Replacing regular batch normalization with its batch-independent analogue involves adding to the standard activation step auxiliary \textit{proxy norm} variable. This variable selectively preserves the expressivity of produced feature map, compensating for the impact of learnable affine transformations and incorporated nonlinearity. 
\\

The results of computational experiments are represented in Figure 7. In some cases, the mentioned techniques have led to a certain decrease in metric values, yet their general influence can be considered positive. Overall, the most significant performance growth was obtained on the images with winter, distortion, and occlusion labels. The combination of techniques that performed best (optimizer, resize, EMA), as well as the contrasting ones (optimizer, grouped convolutions,  batch accumulation), never surpassed the metrics of individual methods. According to the categories' size, the exponential moving average enabled EfficientNet-P to achieve the best quality of predictions. Guided by the provided heatmap, the proper technique has to be chosen on a case-by-case basis, as the applied scenario can be strongly skewed towards certain vision conditions.

\subsection*{\textbf{Inference time analysis}}

In order to evaluate the practical applicability of the proposed architecture, it is required to assess its computational cost and compare it with the same property of the baseline models. As it can be seen from Table~\ref{table:comp_cost}, inference time is proportional to the model size, with the default EfficientNet-P occupying the end of the ranked list.

\begin{table}[t]
    \renewcommand{\arraystretch}{1.5}
    \centering
    \caption{Inference time properties of the CNN classifiers (images per second)}
    \smallskip
    \begin{tabular}{l|l|l|l|l|l|l|l}
    \toprule
         & mAlexNet & AlexNet & VGG-19 & MobileNet & CFEN & ResNet50 & EfficientNet-P \\ 
        \midrule
        Mean & 1242.07 & 972.62 & 319.31 & 234.35 & 227.28 & 185.12 & 87.072 \\
        \midrule
        Std & 36.23 & 25.92 & 10.48 & 11.97 & 18.33 & 9.81 & 2.44 \\
        \bottomrule
    \end{tabular}
    \label{table:comp_cost}
\end{table}

%
%
%
%
%
%

To optimize the computational cost of EfficientNet-P for real-world parking systems, we applied a classic distillation strategy implying teacher-student knowledge transfer~\cite{hinton2014distilling}. Following this concept, the teacher model was represented by the standard version of the proposed pipeline, while the student one corresponded to the truncated EfficientNet-P without several MBConv blocks. These smaller neural networks included 3, 4, 5, and 6 first modules (Figure~\ref{method:pipeline}) of original EfficientNet-B0 with the custom module 8 on top of them. During our experiments, each student model was trained with respect to the altered binary cross-entropy loss function:
\begin{equation}
L = -\frac{1}{N} \sum_{i=0}^N y_i \cdot \log \left(\hat{y}_i\right)+\left(1-y_i\right) \cdot \log \left(1-\hat{y}_i\right) + \alpha\cdot(\boldsymbol{z}_i^{(T)}-\boldsymbol{z}_i^{(S)})^2,
\end{equation}
where $N$ is the number of entities, $y_i$ and $\hat{y}_i$ are ground truth and predicted labels, $\alpha$ is the distillation coefficient fixed at 0.1, $\boldsymbol{z}_i^{(T)}$ and $\boldsymbol{z}_i^{(S)}$ are the logits of the teacher and student models correspondingly.

\begin{figure}[!t]
\centering
\includegraphics[width=348pt]{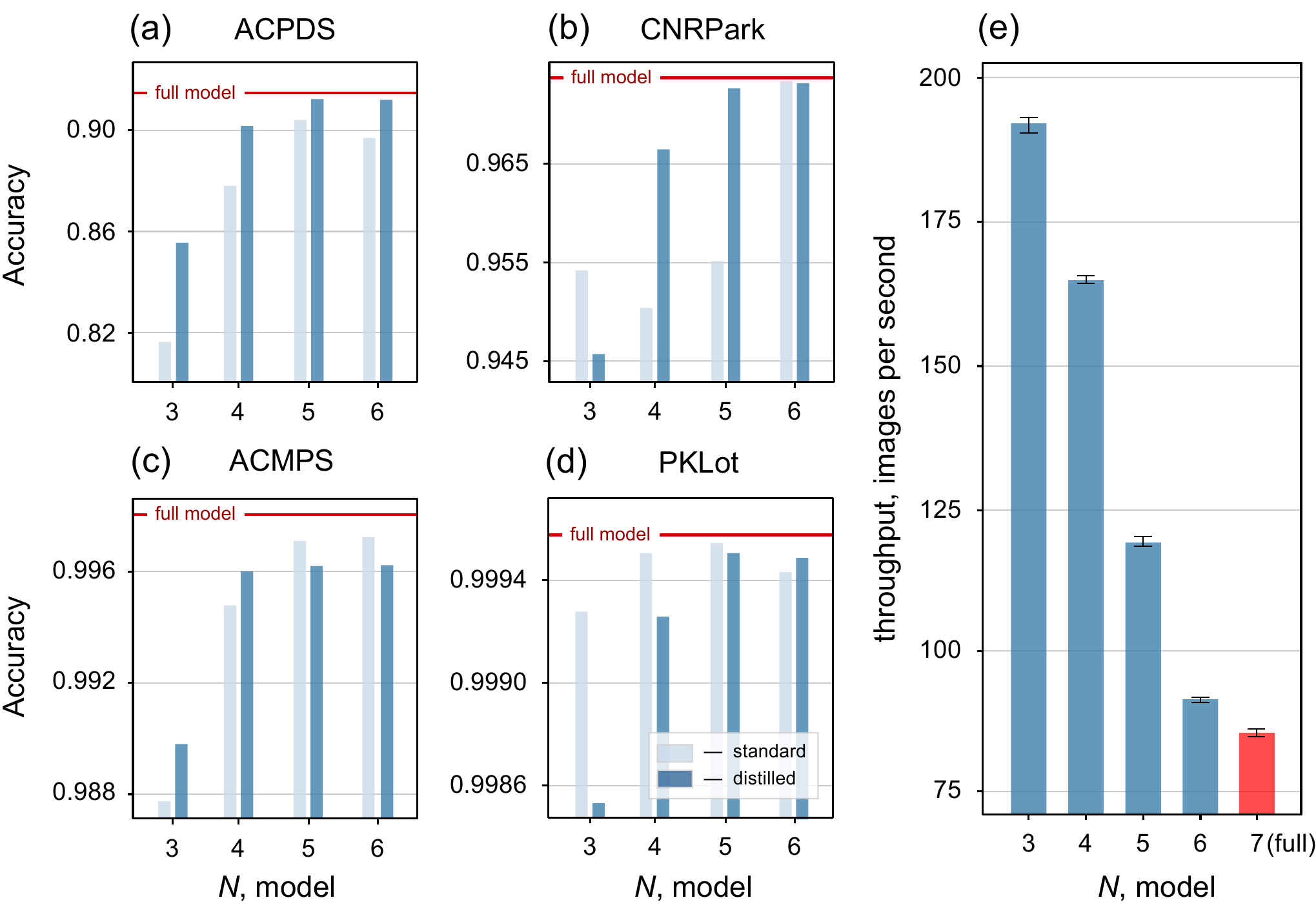}
\caption{
\csentence{Comparison of truncated standard and distilled versions of EfficientNet-P.}
(a, b, c, d) describe prediction quality on corresponding datasets while (e) depicts the throughput of each studied architecture.}
\label{results:result_opt}
\end{figure}

Despite the efficiency of this approach in similar tasks~\cite{touvron2021training, lin2022knowledge}, estimation of the distillation effect is still desirable as it clarifies potential tradeoffs between prediction quality and inference time for considered task. For that purpose, we compared common truncated models with their counterparts based on extended loss (5). Obtained results indicate that the chosen distillation strategy has a positive influence on metric values of student models: for ACPDS and CNRPark,  performance dropdown is observed only for models 3 and 4 while the others have mostly similar results. On the same datasets, the distilled version demonstrated better prediction quality nearly reaching the quality of the original EfficientNet-P. This outcome is compensated to some extent by the worse results on ACMPS and PKlot, where it almost did not surpass the standard approach to training. However, the most crucial property of distilled models is not their quality, but faster convergence with the growth of the modules'  number. Thus, depending on the preference of the end user, one can leverage the chosen distilled model balancing between prediction quality depicted in Figure~\ref{results:result_opt}(a-d) and expected throughput, Figure~\ref{results:result_opt}(e).

\section*{Conclusion}
In this study, we introduced a new predictive algorithm for parking lot occupancy detection and explored its capabilities in different configurations. The performed experiments have shown that the proposed EfficientNet-P model is able to outperform previous solutions while maintaining the same magnitude of computational complexity. Along with the main contribution, a comparison of patch-based and intersection-based approaches has revealed the insufficient performance of the models relying on the object detection algorithms. Additionally, we demonstrated that the current amount of collected data does not allow vision transformers to reach the same performance as classic CNNs regarding parking lot occupancy detection. 

 Performed extension of existing datasets enables to apply them further as the first integral benchmark in the domain of parking lot occupancy detection. We look forward that developed widgets will be actively used in the following studies as the main annotation tools associated with the considered task. 

\section*{Future work}
We consider the presented contributions as the initial step towards consistent evaluation in the area of parking lot occupancy detection. In further studies, we aim to extend our current results as follows:  \begin{enumerate}
    \item Finalize visual conditions labelling for the existing image datasets.
    \item Perform the large-scale testing of image augmentation techniques and empirically derive the most efficient configuration.
    \item Explore the possibility of inference time reduction regarding the best models identified in this study.
\end{enumerate}
In addition, we believe that it will be extra beneficial to design a high-level API based on a client-server architecture. Such an application will tangibly simplify real-world approbation of different predictive models, which is crucial for practitioners involved in the implementation of parking guidance systems.

\newpage

\section*{Supplementary}

\subsection*{S1 -- Separate training \& separate evaluation of patch-based models}

\begin{table}[htp!]
     \renewcommand{\arraystretch}{1.5}
    
    \centering
    \caption{F1-score for patch-based methods (vision transformer classifiers)}
    \smallskip
    \begin{tabularx}{355px}{XX|X|X|X|X|X}
        \toprule
        \multicolumn{2}{l|}{} & ACMPS & ACPDS & CNRPark & PKLot & SPKL \\ 
        \midrule
        \multicolumn{2}{l|}{ViT (no prt)} & 0.9806 & 0.8966 & \textbf{0.9881} & 0.9844 & 0.8727\\ 
        \multicolumn{2}{l|}{ViT (prt)} & 0.9595 & \textbf{0.9370} & 0.9845 & \textbf{0.9983} & \textbf{0.9383} \\ 
        \multicolumn{2}{l|}{DeiT (prt)} & 0.9542 & 0.9201 & 0.9833 & 0.9880 & 0.9027 \\ 
        \multicolumn{2}{l|}{PiT (prt)} & 0.9294 & 0.9372 & 0.9451 & 0.9740 & 0.9333 \\ 
        \multicolumn{2}{l|}{ViT (SPT LSA)} & \textbf{0.9935} & 0.9073 & 0.9804 & 0.9967 & 0.9320 \\ 
        \bottomrule
    \end{tabularx}

    \label{supp_table:vt}
\end{table}

\begin{table}[htp!]
    \renewcommand{\arraystretch}{1.5}
    
    \centering
    \caption{F1-score for patch-based methods (convlutional neural network classifiers)}
    \smallskip
    \begin{tabularx}{355px}{XX|X|X|X|X|X}
        \toprule
        \multicolumn{2}{l|}{} & ACMPS & ACPDS & CNRPark & PKLot & SPKL \\
        \midrule
        \multicolumn{2}{l|}{ResNet50} & 0.9968 & 0.9359 & 0.9798 & 0.9955 & 0.8883 \\
        \multicolumn{2}{l|}{MobileNet} & 0.9992 & 0.9668 & 0.9974 & 0.9995 & 0.9701 \\ 
        \multicolumn{2}{l|}{CarNet} & 0.9938 & 0.8765 & 0.9888 & 0.9975 & 0.9542 \\ 
        \multicolumn{2}{l|}{AlexNet} & 0.9964 & 0.9308 & 0.9920 & 0.9994 & 0.9379 \\ 
        \multicolumn{2}{l|}{mAlexNet} & 0.9882 & 0.8999 & 0.9870 & 0.9962 & 0.9156 \\ 
        \multicolumn{2}{l|}{VGG-16} & 0.9929 & 0.9346 & 0.9933 & 0.9990 & 0.9343 \\ 
        \multicolumn{2}{l|}{VGG-19} & 0.9941 & 0.9593 & 0.9917 & 0.9991 & 0.9222 \\ 
        \multicolumn{2}{l|}{CFEN} & 0.9886 & 0.9098 & 0.9845 & 0.9971 & 0.9156 \\ 
        \midrule
        \multicolumn{2}{l|}{EfficientNet-B0} & 0.9952 & \textbf{0.9950} & 0.9952 & 0.9954 & \textbf{0.9795} \\ 
        \multicolumn{2}{l|}{EfficientNet-P} &\textbf{0.9992} & 0.9866 & \textbf{0.9991} & \textbf{0.9995} & \textbf{0.9795} \\ 
        \bottomrule
        
    \end{tabularx}

    \label{supp_table:cnn}
\end{table}

As well as for the models trained on merged data, for the separate case EfficientNet-P was also able to outperform baselines (on the same subset of the datasets). Vision transformers did not achieve comparable prediction quality with CNN on ACPDS and SPKL while retaining decent results on the rest of the datasets. Overall, metrics appeared to be greater for the separate training setting than for the merged one. This effect can be partially explained by the inner homogeneity of each dataset regarding specific points of view and visual conditions. 

\begin{figure}[!t]
\centering
\includegraphics[width=348pt]{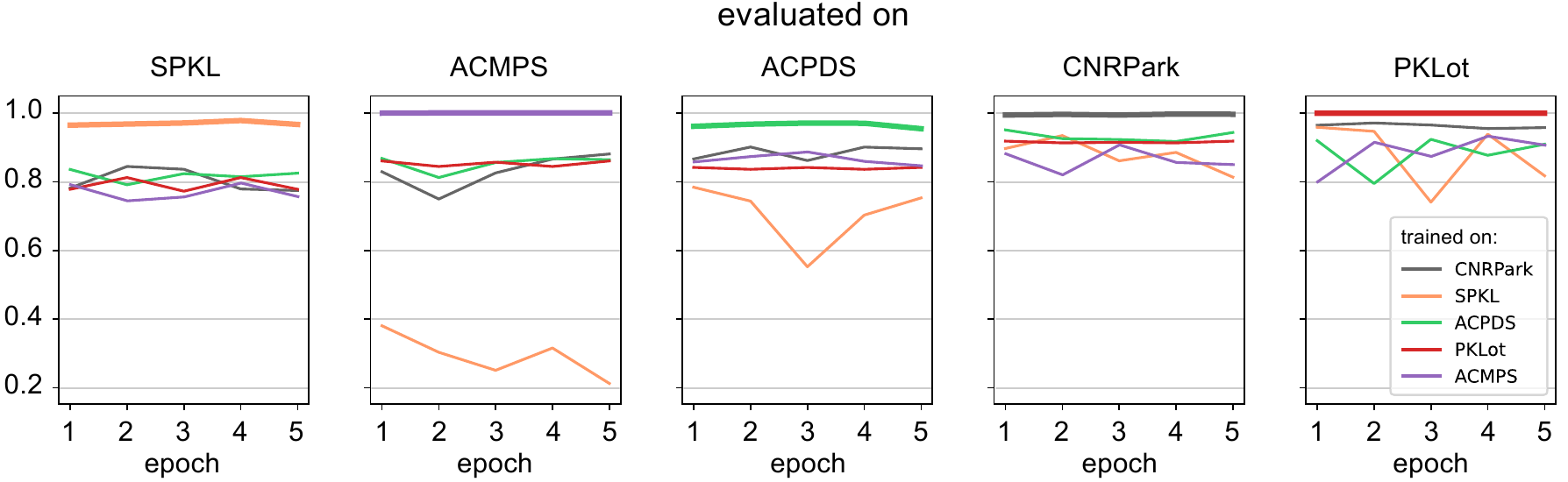}
\caption{
\csentence{Cross-evaluation on 5 datasets}. The figure demonstrates changes in the F1-score for the following 5 epochs after model convergence on the training dataset (bold line) with respect to the validation part of each dataset.}
\label{results:cross_eval_en_p}
\end{figure}

Since the separate and merged approach to training resulted in tangibly different metrics, we performed an additional assessment of the generalization ability on unseen datasets during the training, Figure~\ref{results:cross_eval_en_p}. The cross-evaluation on EfficientNet-P has shown that some datasets tend to be harder to generalize: SPKL does not seem to be the right option as the only data source for training since the model achieved remarkably worse validation results on ACMPS and ACPDS in comparison with the other datasets. In the reversed setting, evaluation on the SPKL validation part yielded the lowest F1-score among the models trained on the other datasets, thereby emphasizing its distinct nature. The most stable dynamics of metrics across the epochs is observed on the CNRPark dataset which appeared to be less prone to overfitting in comparison with ACPDS and ACMPS. Meanwhile, despite being one of the largest benchmarks for parking lot occupancy detection, ACMPS has as low a generalization ability regarding EfficientNet-P as some of the smaller datasets.

\newpage

\subsection*{\textbf{Declarations}}
\vspace{-12pt}

\begin{backmatter}

\section*{Ethics approval and consent to participate}
Not applicable.

\section*{Consent for publication}
Not applicable.

\section*{Availability of data and materials}
Considered models and datasets are available in the project's GitHub repository.

\section*{Competing interests}
The authors declare that they have no competing interests.

\section*{Funding}
Not applicable.

\section*{Authors contributions}
A.M., M.K., and V.T.: Software, Data curation, Validation, Visualization; 
V.P.: Software, Methodology, Writing (original draft), Visualization, Research management; 
A.K.: Conceptualization, Writing (review \& editing);
N.S.: Supervision, Methodology, Writing (review \& editing);
K.K.: Supervision, Writing (review \& editing).

\section*{Acknowledgements}
We are grateful to \anonymize{Daria Petrova} and \anonymize{Daniel Efimov} for their contribution towards the project management.


\bibliographystyle{bmc-mathphys} 
\bibliography{sections/literature.bib}      

\end{backmatter}
\end{document}